\documentclass[11pt]{article}
\usepackage{coling2018}
\usepackage{times}
\usepackage{url}
\usepackage{latexsym}
\usepackage[]{todonotes}
\usepackage{eurosym}
\usepackage{enumitem}
\setitemize{noitemsep,topsep=0pt,parsep=0pt,partopsep=0pt}
\usepackage[utf8x]{inputenc}
\usepackage[greek,english]{babel}
\slname{Greek}

\title{Automated Fact Checking:\\ Task formulations, methods and future directions}
\sltitle{\textgreek{Αναγνώριση Ψευδών Ισχυρισμών:\\ Ορισμοί, μέθοδοι και κατευθύνσεις για μελλοντική έρευνα}}

\author{James Thorne \\
  Department of Computer Science \\
  University of Sheffield, UK \\
  {\tt j.thorne@sheffield.ac.uk} \\\And
  Andreas Vlachos \\
  Department of Computer Science \\
  University of Sheffield, UK \\
  {\tt a.vlachos@sheffield.ac.uk} \\}

\date{}

\begin{document}
\maketitle
\begin{abstract}
The recently increased focus on misinformation has stimulated research  in fact checking, the task of assessing the truthfulness of a claim. Research in automating this task has been conducted in a variety of disciplines including natural language processing, machine learning, knowledge representation, databases, and journalism. While there has been substantial progress, relevant papers and articles have been published in research communities that are often unaware of each other and use inconsistent terminology, thus impeding understanding and further progress.
In this paper we survey automated fact checking research stemming from natural language processing and related disciplines, unifying the task formulations and methodologies across papers and authors. 
Furthermore, we highlight the use of evidence as an important distinguishing factor among them
cutting across task formulations and methods. We conclude with proposing avenues for future NLP research on automated fact checking. 
\end{abstract}

\makesltitle
\begin{slabstract}
\textgreek{Το πρόσφατα αυξημένο ενδιαφέρον για το φαινόμενο της παραπληροφόρησης έχει κινητοποιήσει ερευνητικές προσπάθειες για μεθόδους αναγνώρισης ψευδών ισχυρισμών. Η έρευνα σε αυτοματοποιημένες μεθόδους για το πρόβλημα αυτό διεξάγεται σε διάφορους επιστημονικούς κλάδους, όπως επεξεργασία φυσικής γλώσσας, μηχανική μάθηση, αναπαράσταση γνώσης, βάσεις δεδομένων και δημοσιογραφία. Αν και έχει γίνει σημαντική πρόοδος, τα άρθρα σχετικά με το αντικείμενο δημοσιεύονται στα συνέδρια και τα περιοδικά του κάθε κλάδου και χρησιμοποιούν διαφορετική ορολογία, με συνέπεια να δυσχεραίνεται η περαιτέρω πρόοδος. Στο παρόν άρθρο παρουσιάζουμε μια ανασκόπηση της έρευνας στον αυτόματη αναγνώριση ψευδών ισχυρισμών με κεντρικό άξονα την επεξεργασίας φυσικής γλώσσας, ενοποιώντας ορισμούς και  μεθόδους που έχουν προταθεί στη βιβλιογραφία. Μια βασική διάκριση που διαχωρίζει τις ορισμούς και μεθόδους που έχουν προταθεί σε διαφορετικούς κλάδους είναι η χρήση ή μη αποδεικτικών στοιχείων. Η ανασκόπηση στο άρθρο αυτό καταλήγει με κατευθύνσεις για μελλοντική έρευνα στην επεξεργασία φυσικής γλώσσας σχετική με το πρόβλημα αυτό.}
\end{slabstract}


\blfootnote{This work is licenced under a Creative Commons Attribution 4.0 International Licence. Licence details: \url{http://creativecommons.org/licenses/by/4.0/}}

\section{Introduction}
The ability to 
rapidly distribute information on the internet presents a great opportunity for individuals and organizations to disseminate and consume large amounts of data on almost any subject matter. As many of the barriers required to publish information are effectively removed through the advent of social media, it is possible for content written by any individual or organization
to reach 
an audience of hundreds of millions of readers \cite{allcott2017}. This enhanced ability of reaching audiences can be used to spread both true and false information, and has become an important concern as it is speculated that this has affected decisions such as public votes, especially since recent studies have shown that false information reached greater audiences \cite{Vosoughi1146}. 
Consequently, there has been an increased demand for fact checking of online content.


In the domain of journalism, fact checking is the task of assessing whether claims made in written or spoken language are true. 
This is a task that is normally performed by trained professionals: the fact checker must evaluate previous speeches, debates, legislation and published figures or known facts, and use these, combined with reasoning to reach a verdict. 
Depending on the complexity of the claim, this process may take from less than one hour to a few days \cite{Hassan2015}. 
Due to the large volume of content to be checked, this process has been the subject of calls from the journalism community to develop tooling to automate parts of this task \cite{Cohen2011,Babakar2016,Graves2018AFC}. 
The process of fact checking requires researching and identifying evidence, understanding the context of information and reasoning about what can be inferred from this evidence. The goal of automated fact checking is to reduce the human burden in assessing the veracity of a claim.

In this paper we survey automated fact checking work from the viewpoint of natural language processing (NLP) and related fields such as machine learning, knowledge representation, databases and social media analysis. 
While it is important to consider social aspects, including how fact checking is communicated and incorporated into social media platforms, these are considered out of scope for this survey. 
There is a vast body of related works with different motivations and approaches that we can learn from to develop natural language technologies for fact checking. 
However, because these are conducted in a siloed manner and often without wider awareness of the issue, there is inconsistency in the definitions and lack of communication between disciplines, impeding understanding and further progress.
For example, the classification of whether an article title can be supported by content in the body has been interchangeably referred to as both fake news detection \cite{FNC}, stance classification \cite{Ferreira2016} and incongruent headline detection \cite{Chesney2017}. 

In this survey we aim to unify the definitions presented in related works and identify common concepts,
datasets and modelling approaches. 
In this process, we summarize key results, highlight limitations and propose some open research challenges. 
In particular, we detail the different types of evidence used by different approaches and how they affect the NLP requirements on the systems developed.
For example, many approaches rely on textual entailment/natural language inference \cite{Dagan2009}, while others rely on knowledge base construction techniques \cite{Ji:2011:KBP:2002472.2002618}. 




The structure of this survey is as follows: we first discuss fact checking in the context of journalism as this provides definitions and distinctions on key terminology that will be used throughout in the remainder. We then proceed to discussing previous research in automated fact checking in terms of what inputs they expect, what outputs they return and the evidence used in this process. Following this, we provide an overview of the most commonly used datasets and the models developed and evaluated on them. Subsequently, we discuss work related to automated fact checking and conclude by proposing avenues for future NLP research. 

\section{Fact checking in journalism}
Fact checking is an essential component in the process of news reporting.
Journalism has been defined by scholars as a ``discipline of verification'' to separate it from ``entertainment, propaganda, fiction, or art'' \cite{doi:10.1080/17512786.2013.765638}.
While fact checking and verification are often used interchangeably, recent work has sought to define them as two distinct but complementary processes \cite{Silverman2016verification}. 
In particular, verification is defined as ``scientific-like approach of getting the fact and also the right facts'' \cite{kovach2014elements}, which often involves verifying the source, date and the location of materials. 
Fact-checking on the other hand ``addresses the claim's logic, coherence and context'' \cite{Mantzarlis}. 
Consequently, verification is a necessary and crucial first step in the process of fact checking as it assesses the trustworthiness of the contexts considered. 
We adopt this distinction for the remainder of this survey as it helps highlight key differences between various automated approaches.\footnote{Note that Mantzarlis and Silverman disagree on whether fact-checking should be considered a subset of verification, or just overlapping with it. While this is in an important question, we do not address it in this survey as it does not affect our analysis.}

A term that has become strongly associated with fact checking is ``fake news'', especially since its use in the context of the 2016 US presidential elections. 
However its popularity has resulted in its meaning becoming diluted, as it is used to label claims on a number of aspects not necessarily related to veracity \cite{Vosoughi1146}. 
The most prominent example of such usage of the term of fake news is its application to media organizations of opposing political sides. Furthermore, it is often grouped together with the term ``hate speech'' as in the case of recent legislation \cite{dw2017}, despite the latter being more related to the use of emotive language instead of truth assessment \cite{Rahman:2012}. Therefore we avoid using this term in this survey.

A further consideration is the relation of fact checking with the terms misinformation and disinformation. 
While the former is the 
distribution of information that may not be accurate or complete, the
latter additionally assumes 
malicious motives to mislead the reader \cite{jowett2006propaganda}. 
Due to this additional consideration, disinformation is considered a subset of misinformation. 
Fact checking can help detect misinformation, but not distinguish it from disinformation. 





\section{Towards Automated Fact Checking}
The increased demand for fact checking has stimulated a rapid progress in developing tools and systems to automate the task, or parts thereof \cite{Babakar2016,Graves2018AFC}. 
Thus, there has been a diverse array of approaches 
tailored to specific datasets, domains or subtasks. 
While they all share the same goal, these approaches utilize different definitions of what the task being automated is.
In the following discussion, we highlight the differences between the task definitions used in previous research in the following axes:
input, i.e.\ what is being fact checked, output, i.e.\ what kinds of verdicts are expected, and the evidence used in the fact checking process. 


\subsection{Inputs}
We first consider the inputs to automated fact checking approaches as their format and content influences the types of evidence used in this process.
A frequently considered input to fact checking approaches is 
subject-predicate-object triples, e.g. {\tt (London, capital\_of, UK)}, and is popular across different research communities, including NLP \cite{Nakashole2014}, data mining \cite{Ciampaglia2015} and Web search \cite{bast2015relevance}. The popularity of triples as input stems from the fact that they facilitate fact checking against (semi-)structured knowledge bases such Freebase \cite{Bollacker2008}. However, it is important to acknowledge that approaches using triples as input implicitly assume a non-trivial level of processing in order to convert text, speech or other forms of claims into triples, a task falling under the broad definition of natural language understanding \cite{Woods:1973:PNL:1499586.1499695}.

A second type of input often considered in automated fact checking is textual claims. These tend to be short sentences constructed from longer passages, which is a practice common among human fact checkers on dedicated websites such as PolitiFact\footnote{\url{http://www.politifact.com/}}
and Full Fact\footnote{\url{http://www.fullfact.org/}} with the purpose of including only the context relevant to the claim from the original passage. The availability of fact checked claims on such websites has rendered this format very popular among researchers in NLP.


A useful taxonomy of textual claims was proposed in the context of the HeroX fact checking challenge \cite{fullfact2016herox}, in which four types of claims were considered:
\\
\begin{itemize}
\item \textbf{numerical claims} involving numerical properties about entities and comparisons among them
\item \textbf{entity and event properties} such as professional qualifications and event participants
\item \textbf{position statements} such as whether a political entity supported a certain policy 
\item \textbf{quote verification} assessing whether the claim states precisely the source of a quote, its content, and the event at which it supposedly occurred.
\end{itemize}
\clearpage
While some of these claims could be represented as triples, they typically require more complex representations; for example, events typically need to be represented with multiple slots \cite{ACE04} to denote their various participants. Regardless of whether textual claims are verified via a subject-predicate-object triple representation or as text, it is often necessary to disambiguate the entities and their properties. For example, the claim `Beckham played for Manchester United' is true for the soccer player `David Beckham', but not true (at the time of writing) for the American football player `Odel Beckham Jr'. Correctly identifying, disambiguating and grounding entities is the task of Named Entity Linking \cite{mcnamee2009}. While this must be performed explicitly if converting a claim to a subject-predicate-object triple with reference to a knowledge base, it may also be performed implicitly through the retrieval of appropriate textual evidence if fact checking against textual sources.


Finally, there have been approaches that consider an entire document as their input. These approaches must first identify the claims and then fact check them.
This increases the complexity of the task, as approaches are required to extract the claims, either in the form of triples by performing relation extraction \cite{Vlachos2015} or through a supervised sentence-level classification \cite{Hassan2015a}.


\subsection{Sources of evidence}


The type of evidence that is used for fact checking influences the model and the types of outputs that the fact checking system can produce. 
For example, whether the output is a 
label or whether a justification can be produced depends largely on the information available to the fact checking system.

We first consider 
task formulations
that do not use any evidence beyond the claim itself when predicting its veracity such as the one by \newcite{Rashkin2017}. 
In these instances, surface-level linguistic features in the claims are associated with the predicted veracity. We contrast this to how journalists work when fact checking, where they must find knowledge relating to the fact and evaluate the claim given the evidence and context when making the decision as to whether a claim is true or false. 
The predictions made in task formulations that do not consider evidence beyond the claim are based on surface patterns of how the claim is written rather than considering the current state of the world.

\newcite{Wang2017a} incorporate additional metadata in fact checking such as the originator of the claim, 
speaker profile and the media source in which the claim is presented. While these do not provide evidence grounding the claim, the additional context can act as a prior to improve the classification accuracy, and can be used as part of the justification of a verdict. 

Knowledge graphs provide a rich collection of structured canonical information about the world stored in a machine readable format that could support the task of fact checking. We observe two types of formulations using this type of evidence. 
The first approach is to identify/retrieve the element in the knowledge graph that provides the information supporting or refuting the claim at question. For example,
 \newcite{Vlachos2015} and \newcite{Thorne2017}
identify the subject-predicate-object triples from small knowledge graphs to fact check numerical claims. 
Once the relevant triple had been found, a truth label is computed through a rule based approach that considers the error between the claimed values and the retrieved values from the graph. 
The key limitation in using knowledge graphs as evidence in this fashion is that it assumes that the true facts relevant to the claim are present in them. 
However, it is not feasible to capture and store every conceivable fact in the graph in advance of knowing the claim. 


The alternative use of a knowledge graph as evidence is to consider its topology in order to predict how likely a claim (expressed as an edge in the graph) is to be true \cite{Ciampaglia2015}. 
While graph topology can be indicative of the plausibility of a fact, nevertheless if a fact is unlikely to occur that does not negate its truthfulness.
Furthermore, improbable but believable claims are more likely to become viral and thus in greater need of verification by fact checkers.

Text, such as encyclopedia articles, policy documents, verified news and scientific journals contain information that can be used to fact check claims. \newcite{Ferreira2016} use article headlines (single sentences) as evidence to predict whether an article is for, against or observing a claim.
The Fake News Challenge \cite{FNC} also formulated this part of the fact checking process in the same way, but in contrast to \cite{Ferreira2016}, entire documents are used as evidence, thus allowing for evidence from multiple sentences to be combined. 

The Fact Extraction and VERification (FEVER) task \cite{Thorne2018} requires combining information from multiple documents and sentences for fact checking. 
Unlike the aforementioned works which use text as evidence, the evidence is not given but must be retrieved from Wikipedia, a large corpus of encyclopedic articles. 
Scaling up even further, the triple scoring task of the WSDM cup \cite{Bast2017} required participants to assess knowledge graph triples considering both Wikipedia and a portion of the web-scale ClueWeb dataset \cite{FACC1}. 

A different source of text-based evidence is repositories of  previously fact checked claims \cite{Hassan2017}.
Systems using such evidence typically match a new claim to claim(s) in such a repository and return the label of the match if one is found. 
This type of evidence enables the prediction of veracity labels instead of only whether a claim is supported or refuted. However, this evidence is limiting fact checking only to claims similar to the ones already existing in the repository. 

As an alternative to using knowledge stored in texts or databases as evidence, aggregate information on the distribution of posts on social networks can provide insights into the veracity of the content. Rumor veracity prediction is the assessment of the macro level behaviors of users' interactions with and distribution of content to predict whether the claims in the content are true or false \cite{Derczynski2015,Derczynski2017}. This crowd behavior provide useful insights, especially in cases where textual sources or structured knowledge bases may be unavailable.

The trustworthiness of the sources used as evidence, i.e.\ the verification aspect of fact checking in journalistic terms, is rarely considered by automated approaches, despite its obvious importance. Often the sources of evidence are considered given, e.g. Wikipedia or Freebase, as this facilitates development and evaluation, especially when multiple models are considered. Nevertheless, approaches relying on Twitter metadata often consider credibility indicators for the tweets used as evidence in their fact checking process \cite{Liu:2015:RRD:2806416.2806651}. Similar to journalism, trustworthiness assessment is often considered as a separate task and we discuss it further in the Related Work section.

\subsection{Output}

The simplest model for fact checking is to label a claim as true or false as a binary classification task \cite{Nakashole2014}. 
However, we must also consider that it is possible in natural language to be purposefully flexible with the degree of truthfulness of the information expressed or express a particular bias using true facts. Journalistic fact checking agencies such as Politifact 
model the degree of truthfulness on a multi-point scale (ranging from true, mostly-true, half-true, etc). Rather than modeling fact checking as a binary classification, \newcite{Vlachos2014} suggested modeling this degree of truthfulness as an ordinal classification task. However, the reasoning behind why the manual fact checking agencies have applied these more fine-grained labels is complex, sometimes inconsistent\footnote{\url{http://www.poynter.org/news/can-fact-checkers-agree-what-true-new-study-doesnt-}
\url{point-answer}} and likely to be difficult to capture and express in our models. 
\newcite{Wang2017} and \newcite{Rashkin2017} expect as output multiclass labels following the definitions by the journalists over a multi-point scale but ignoring the ordering among them.

The triple scoring task of the WSDM cup \cite{Bast2017} expected as output triples scored within a numerical range indicating how likely they are to be true. The evaluation consisted of calculating both the differences between the predicted and the manually annotated scores, as well as the correlation between the rankings produced by the systems and the annotators.

\newcite{Ferreira2016} expect as output whether a claim is supported, refuted or just reported by a news article headline. \newcite{FNC} added an extra label for the article being irrelevant to the claim, and consider the full article instead of the headline. While this output can be used in the process of fact checking, it is not a complete fact check.

In FEVER \cite{Thorne2018} the output expected consists of two components: a 3-way classification label about whether a claim is supported/refuted by Wikipedia, or there is not enough information in the latter to Wikipedia it, and in the case of the first two labels,
the sentences forming the evidence to reach the verdict. In effect this combines the multiclass labeling output with a ranking task over Wikipedia sentences. If the label predicted is correct but the evidence retrieved is incorrect, then the answer is considered incorrect, highlighting the importance of evidence in this task formulation. 
Nevertheless, this output cannot be considered a complete fact check though, unless we restrict world knowledge to Wikipedia. The use of full world knowledge in the justifications was considered in the HeroX shared task \cite{fullfact2016herox}; while this allowed for complete fact checks, it also necessitated manual verification of the outputs.




\section{Fact Checking Datasets}

There are currently a limited number of published datasets resources for fact checking.
\newcite{Vlachos2014} collected and released 221 labeled claims in the political domain that had been fact checked by Politifact 
and Channel4.\footnote{\url{http://channel4.com/news/factcheck}} For each labeled claim, the speaker or originator is provided alongside hyperlinks to the sources used by the fact checker as evidence. The sources range from statistical tables and Excel spreadsheets to PDF reports and documents from The National Archives. While these sources can be readily assessed by human fact checkers, the variety and lack of structure in the evidence 
makes
it difficult to apply machine learning-based approaches to select evidence as separate techniques may be required for the each different document format. Where documents are provided as evidence, we do not know which portions of the document pertain to the claim, which compounds the difficulty of assessing and evaluating a fact checking system, given the need for evidence and accountability. Furthermore, with such a limited number of samples, the scale of this dataset precludes its use for developing machine learning-based fact checking systems.

\newcite{Wang2017a} released a dataset similar to but an order of magnitude larger than that of \newcite{Vlachos2014} containing 12.8K labeled claims from Politifact. In addition to the claims, meta-data such as the speaker affiliation and the context in which the claim appears in (e.g.\ speech, tweet, op-ed piece) is provided. At this scale, this dataset can support training and evaluating a machine learning-based fact checking system. However, the usefulness of the dataset may be limited due the claims being provided without machine-readable evidence beyond originator metadata, meaning that systems can only resort to approaches to fact checking such as text classification or speaker profiling. 

\newcite{Rashkin2017} 
collated claims from Politifact without meta-data. In addition, the authors published a dataset of 74K news articles collected from websites deemed as Hoax, Satire, Propaganda and Trusted News. The prediction of whether a news article is true or false is modeled as the task of predicting whether an article originates from one of the websites deemed to be ``fake news'' according to a US News \& World report.\footnote{\url{www.usnews.com/news/national-news/articles/2016-11-14/avoid-these-fake-news-sites-}
\url{at-all-costs}} This allows systems to fact check using linguistic features, but does not consider evidence, or which aspects of a story are true or false.

\newcite{Ferreira2016} released a dataset for rumor debunking using data collected from the Emergent project \cite{Silverman2015}. For 300 claims, 2,595 corresponding news articles were collected and their stances were labeled as for, against or observing a claim. This dataset  was extended for the 2017 Fake News Challenge \cite{FNC} dataset which consisted of approximately 50K headline and body pairs derived from the original 300 claims and 2,595 articles. 

The HeroX Fast and Furious Fact Checking Challenge \cite{fullfact2016herox} 
released 90 (41 practice and 49 test) labeled claims as part of the competition. Because part of the challenge required identification of appropriate source material, evidence for these claims was not provided and therefore manual evaluation was needed on case-by-case basis. As with the claims collected by \newcite{Vlachos2014}, the size of the dataset prevents its use for training a machine learning-based fact checking system. However, the broad range of types of claims in this dataset highlights a number of forms of misinformation to help identify the requirements for fact checking systems.

\newcite{Thorne2018} introduced a fact checking dataset containing 185K claims about properties of entities and concepts which must be verified using articles from Wikipedia. While this restricts the pool of evidence substantially compared to HeroX, it allowed for a challenging evidence selection subtask that could be feasibly annotated manually at a large scale in the form of sentences selected as evidence. Combined with the labels on the claims, these machine readable fact checks allow training machine learning-based fact checking models.



\section{Methods}
The majority of the methods used for automated fact checking are supervised: learning a text classifier from some form of labeled data that is provided at training. 
This trait that is independent of the task input or what sources of evidence are considered. \newcite{Vlachos2014} suggested fact checking using supervised models, making use of existing statements that had been annotated with verdicts by fact checking agencies and journalists. 
\newcite{Wang2017a} and \newcite{Rashkin2017} applied this approach to classify the veracity of fake news stories. 
The main limitation of text classification approaches is that fact checking a claim requires additional world knowledge, typically not provided with the claim itself. 
While language can indicate whether a sentence is factual \cite{Nakashole2014}, credible sounding sentences may also be inherently false.
Text classification on claims alone has been used for the related task of detecting fact-check worthy claims \cite{hassan2017toward}.

\newcite{Ciampaglia2015} use network analysis to predict whether an unobserved triple is likely to appear in a graph by modeling the task as a path ranking problem \cite{Lao2011}. 
The truth verdict is derived from the cost of traversing a path between the two entities under transitive closure, weighted by the degree of connectedness of the path.
\newcite{Nakashole2014} combine linguistic features on the subjectivity of the language used with the co-occurrence of a triple with other triples from the same topic, a form of collective classification.



\newcite{Ferreira2016} modeled fact checking as 
a form of Recognizing Textual Entailment (RTE) \cite{Dagan2009,Bowman2015}, predicting whether a premise, typically (part of) a news article, is \emph{for}, \emph{against}, or \emph{observing} a given claim. The same type of model was used by most of the 50 participating teams in the Fake News Challenge \cite{FNC}, including most of the top entries \cite{riedel2017fnc,hanselowski2017fnc}.

The RTE-based models assume that the textual evidence to fact check a claim is given.
Thus they are inapplicable in cases where this is not provided (as in HeroX), or it is a rather large textual resource such as Wikipedia (as in FEVER). 
For the latter, \newcite{Thorne2018} developed a pipelined approach in which the RTE component is preceded by a document retrieval and a sentence selection component. 
There is also work focusing exclusively on retrieving sentence-level evidence from related documents for a given claim \cite{Hua2017}.

\newcite{Vlachos2015} and \newcite{Thorne2017} use distantly supervised relation extraction \cite{Mintz2009} to identify surface patterns in text which describe relations between two entities in a knowledge graph. 
Because these fact checking approaches only focus on statistical properties of entities, identification of positive training examples is simplified to searching for sentences containing numbers that are approximately equal to the values stored in the graph. 
Extending this approach to entity-entity relations would pose different challenges, as a there may be many relations between the same pair entities that would need to be accurately distinguished for this approach to be used.


A popular type of model often employed by fact checking organizations in their process is that of matching a claim with existing, previously fact checked ones. 
This reduces the task to sentence-level textual similarity as suggested by \newcite{Vlachos2014}
and implemented in ClaimBuster \cite{Hassan2017}, Truthteller by The Washington Post\footnote{\url{https://www.knightfoundation.org/articles/debuting-truth-teller-washington-post-}
\url{real-time-lie-detection-service-your-service-not-quite-yet}} and one of the two modes of Full Fact's Live platform.\footnote{\url{https://fullfact.org/blog/2017/jun/automated-fact-checking-full-fact/}} 
However, it can only be used to fact check repeated or paraphrased claims.

\newcite{Long2017} extend the models produced by \newcite{Wang2017a} and improve accuracy of a simple fact checking system through more extended profiling of the originators of the claims. 
The most influential feature in this model is the \emph{credit history} of the originator, a metric describing how often the originator's claims are classified as false. 
This feature introduces a bias in the model that fits with the adage ``never trust a liar''. 
In the case of previously unannotated sources which may haven no recorded history, this feature would be unavailable. 
Furthermore, the strong reliance on credit history has some important ethical implications that need to be carefully considered. 
Despite these limitations, speaker profiling has been shown to be effective in other related studies \cite{Gottipati2013,Long2016} and is discussed further in Section~\ref{sec:bg:speaker}.

\section{Related Tasks}

\paragraph{Verification}
In Section 2, we highlighted the difference between verification and fact checking. 
While the models considered in Section 4 operate under a closed-world paradigm where we assume that all evidence provided is true, for these fact checking technologies to scale-to and operate on the web, we must also consider information that is of unknown veracity as evidence. 

Methods of predicting the authoritativeness of web pages such as PageRank \cite{Brin1998} only consider the hyperlink topology. TrustRank \cite{Gyongyi2004} provides a framework which incorporates annotated information and predicts the trustworthiness of pages based on graph-connectedness to known-bad nodes rather than the information content. An alternative is Knowledge-based Trustworthiness scoring \cite{Dong2015}, which allows predicting whether the facts extracted from a given document page are likely to be accurate given the method used to extract the facts and the website in which the document is published. 

\paragraph{Common Sense Reasoning} Fact checking requires the ability to reason about arguments with common sense knowledge. This requires developing systems that go beyond recognizing semantic phenomena more complex than those typically considered in textual entailment tasks. \newcite{habernal2018argument} introduced a new task and dataset for predicting which implicit \emph{warrant} (the rationale of an argument) is required to support a claim from a given premise. \newcite{Angeli2014} proposed a method of extracting common sense knowledge from WordNet for reasoning about common sense knowledge using Natural Logic 
and evaluated their approach on a subset of textual entailment problems in the {FraCaS} test suite \cite{Cooper1996}. 
It is important to build systems that can reason about both explicit world knowledge and implicit common sense knowledge is an essential step towards automating fact checking.




\paragraph{Subjectivity and Emotive Language}
\newcite{Rashkin2017} assess the reliability of entire news articles by predicting whether the document originates from a website classified as Hoax, Satire or Propaganda. This work is an instance of subjective language detection and does not represent evidence-based fact checking. The authors used supervised classifiers augmented with lexicons including the Linguistic Inquiry and Word Count (LIWC) \cite{Pennebaker2015}, a sentiment lexicon \cite{Wilson2005}, a hedging lexicon \cite{Hyland2015}, and a novel `dramatic' language lexicon to identify emotive, subjective and sensational language in the article bodies. Furthermore, analysis of the lexical features using a logistic regression classifier shows that the highest weighted (most distinguishing) features for the unreliable sources included the use of hedge words (such as `reportedly') or words pertaining to divisive topics (such as `liberals' or `Trump'). The authors apply a similar model to the prediction of claims made by politicians from claims collected from Politifact. The addition of the LIWC lexicon which provides an indication of emotive tone and authenticity marginally improved the classification accuracy for simple lexical models.  

\paragraph{Deceptive Language Detection}
\todo[disable]{User generated content such as reviews on websites may be intentionally false to aid business or for personal gain. A recent notable example is `The Shed at Dulwich' - a `fake' restaurant on TripAdvisor was ranked \#1 in London \footnote{\url{https://www.theguardian.com/lifeandstyle/2017/dec/17/the-shed-tripadvisors-best-london-restaurant-you-literally-couldnt-get-a-table}}}

There are linguistic cues and features in written text that are useful in identifying deceptive language \cite{Zhou2004}. In the context of detecting deceptive user-generated content -  a specific form of disinformation, \newcite{Mihalcea2009} use a simple lexical classification model without further feature engineering. Analysis of the model identifies a number of word classes of the LIWC lexicon which pertain  only to the deceptive texts. \newcite{Ott2011} incorporate the use of psycholinguistic cues to improve classification accuracy. \newcite{Mihalcea2009} found that truthful texts were more likely to contain words belonging to the `optimistic' LIWC class such as `best', `hope', and `determined'. This is corroborated by the study of sentiment in deceptive texts \cite{Ott2013} which also identified that texts with negative sentiment were more likely to be deceptive. These feature classes however may be an artifact of the data generation process as crowd-sourced volunteers were first asked to write an honest text and rewrite it to make it deceptive. 


\newcite{Feng2012} detect deceptive texts and customer-generated reviews through the use of syntactic style rather word-based content. Non-terminal nodes of the constituency parse trees are used as features in conjunction with a lexical model to increase the accuracy over using words alone. \newcite{Hai2016} identify deceptive reviews also using lexical features. However, rather than relying on labeled data, the authors induce labels over an unlabeled dataset through a semi-supervised learning approach that exploits a minimal amount labeled data from related tasks in a multi-task learning set up.

Even though linguistic content, emotive language and syntax are useful indicators for detecting deceit,  the truthfulness of a statement depends also on the context. 
Without considering these factors these approaches cannot be used to fact check information alone.

\paragraph{Rumor Detection}
Rumor detection \cite{Qazvinian2011} 
is the task of identifying unverified reports circulating on social media. A prediction is typically based on language subjectivity and growth of readership through a social network. While these are important factors to consider, a sentence can be true or false regardless of whether it is a rumor \cite{Zubiaga:2018:DRR:3186333.3161603}. 

\paragraph{Speaker Profiling}
\label{sec:bg:speaker}
Identifying claims that do not fit with the profile of the originator may provide insights as to whether the information is truthful or not. Furthermore, determining which topics the originator is truthful about may allow for generation of a risk-based approach to fact checking.
As mentioned earlier, \newcite{Long2017} introduced a notion of credit history which increases the classification accuracy for fake news detection. However, this notion doesn't account for which topics the originator lies about. Furthermore, the assumption that each source has an overall trustworthiness score to be attached to every claim from there
is not a valid one, since inaccurate information may be found even on the most reputable of sources. 

An alternative is to consider the compatibility of a claim with respect to the originator's profile. This remains an open research area. \newcite{Feng2013} perform the inverse of this task for deceptive language detection in product reviews by creating an average profile for the targets (products) and using the distance between a review and the target as a feature. The shortcomings of this method are that number of reviews are required for each target. Considering the tasks of verifying automatically extracted information or fact checking for politics, for a new topic the challenge is that there may be insufficient data to create a profile for it.

\newcite{Perez-Rosas2015} identify author characteristics (such as age and gender) that influence the linguistic choices made by the authors when fabricating information in product reviews. As the affiliation, age and gender of most politicians is public-domain knowledge, it is conceivable that these features may assist fact checking political claims. While the use of meta-data does improve classification accuracy \cite{Wang2017,Long2017}, for fact checking we must consider the meaning of the claim with respect to ground truth rather than based on the linguistic style of the originator.

\paragraph{Click Bait Detection} Intentionally misleading headlines or titles that are designed specifically to encourage a user to click through and visit a website are called click bait. Studies into the detection of click bait have have yielded positive results from relatively simple linguistic features \cite{Chen2015,Potthast2016,Chakraborty2016}. These approaches only consider the article headline and do not make use of evidence. We contrast this to the task of detecting headlines which are incongruent to the document body \cite{Chesney2017}, where existing methods for recognizing textual entailment such as those used for the Fake News Challenge \cite{FNC} can be applied.






\section{Open Research Challenges}
\todo[disable]{Outlook: Does the paper identify areas for future work and/or clearly point out what is not yet handled within the literature surveyed?
Context: Does the paper situate current research appropriately within its historical context? (We don’t expect papers to start with Pāṇini, yet at the same time something that only cites work from 2017 probably doesn’t capture how current work relates to the bigger picture.)}

In this paper, we provided an NLP-motivated overview 
of fact checking, considering the case for evidence - similar to how the task is performed by journalists. We highlighted existing works that consider fact checking of claims expressed in text or via knowledge base triples and pointed out their shortcomings, given the need to justify their decisions using evidence. 

We did not identify approaches that make use of open-world knowledge. Assessing the ability of a system in an open-world setting will be difficult; the HeroX challenge resorted to fully manual evaluation, which is difficult to scale up, especially for the purposes of developing machine learning-based approaches.
Thus we need to consider how to address the information retrieval challenge of the task, including its evaluation.
We must also consider the verification of the evidence used, which is ignored under the closed world assumption. Ideally we should consider verification jointly with fact checking, which is in fact how it is conducted in journalism.

The relative scarcity of resources designed explicitly for fact checking, 
highlights the difficulties in capturing and formalizing the task, especially considering the strong interest in it. While recently published large scale resources such as FEVER can stimulate progress, they only
consider simple short sentences (8 words long on average). Thus, there is scope to fact check compound information or complex sentences, and scale up to fact checking at the document level.

Furthermore, in FEVER the justification for the labels is restricted to sentences selected from Wikipedia. This is much unlike the rationales produced by human fact checkers, who synthesize information. The generation of such rationales has attracted attention only recently in NLP \cite{ling2017program}, and automated fact checking could provide an ideal testbed to develop it further.

While text is often used to make a claim, often the evidence need for fact checking appears in other modalities, such as images and videos. Given the recent interest in multi-modal NLP, we argue that this would be an important direction for future research, especially when considering that a lot of the verification efforts in journalism are focused on identifying forged images and footage.

Finally, it is important to acknowledge that the complexity of the fact checking conducted by journalists is for the moment beyond the abilities of the systems we can develop due to the rather complex reasoning needed. Even a simple short statement such as that the ``UK, by leaving the EU, will take back control of roughly £350 million per week'' takes a substantial amount of work to be checked.\footnote{\url{https://fullfact.org/europe/350-million-week-boris-johnson-statistics-authority}\\\url{-misuse/}} In this fact check for example, the first step is to adjust the claim to render it more accurate in terms of its meaning so that the actual fact check can proceed. While this complexity can help stimulate progress in NLP and related fields, it should also calibrate our expectations and promises to society.


\section*{Acknowledgements}

Andreas Vlachos is supported by the EU H2020 SUMMA project (grant agreement number 688139). The authors would like to thank Dhruv Ghulati and Alexios Mantzarlis for their feedback.

\bibliographystyle{acl}
\bibliography{bib}

\end{document}